\documentclass{article}

\usepackage[preprint]{neurips_2023}

\usepackage[utf8]{inputenc} %
\usepackage[hidelinks,colorlinks=true,linkcolor=blue,citecolor=blue]{hyperref}       %
\usepackage[T1]{fontenc}    %
\usepackage{hyperref}       %
\usepackage{url}            %
\usepackage{booktabs}       %
\usepackage{amsfonts}       %
\usepackage{nicefrac}       %
\usepackage{microtype}      %
\usepackage{xcolor}         %

\title{Policy-Gradient Training of
Language Models \\ for Ranking}

\author{
Ge Gao \hspace{1.1em} Jonathan D. Chang  \hspace{1.1em} Claire Cardie \hspace{1.1em}  Kiant\'e Brantley \hspace{1.1em} Thorsten Joachims \\
\hspace{0.05em} Department of Computer Science, Cornell University \\
\hspace{0.05em} \texttt{ggao@cs.cornell.edu}
\hspace{1.2em} \texttt{\{jdc396,ctc9,kdb82\}@cornell.edu} 
\hspace{1.2em} \texttt{tj@cs.cornell.edu} 
}

\usepackage{xcolor,colortbl}
\usepackage{hyperref}
\usepackage{algorithm, algorithmic}
\usepackage{bm} 
\usepackage{booktabs}
\usepackage{pifont}
\usepackage{multirow}
\usepackage{graphicx}

\usepackage{times}
\usepackage{latexsym}
\usepackage{amssymb}
\usepackage{amsmath}
\usepackage{cleveref}

\usepackage{inconsolata}
\usepackage{microtype}

\usepackage{url}

\newcommand{\alg}{Neural PG-RANK}

\newcommand{\rel}{\mathop{\mathrm{rel}}}
\newcommand{\E}{\mathbb{E}}

\definecolor{redberry}{HTML}{cc4125} %
\definecolor{lightgreen}{HTML}{d9ead3} %

\DeclareMathOperator*{\argmax}{argmax}

\DeclareMathOperator*{\argsort}{argsort}

\begin{document}

\maketitle

\vspace{5pt}
\begin{abstract}
Text retrieval plays a crucial role in incorporating factual knowledge for decision making into language processing pipelines, ranging from chat-based web search to question answering systems. 
Current state-of-the-art text retrieval models leverage pre-trained large language models (LLMs) to achieve competitive performance, but training LLM-based retrievers via typical contrastive losses requires intricate heuristics, including selecting hard negatives and using additional supervision as learning signals.
This reliance on heuristics stems from the fact that the contrastive loss itself is heuristic and does not directly optimize the downstream metrics of decision quality at the end of the processing pipeline. To address this issue, we introduce \alg{}, a novel training algorithm that learns to rank by instantiating a LLM as a Plackett-Luce ranking policy. \alg{} provides a principled method for end-to-end training of retrieval models as part of larger decision systems via policy gradient, with little reliance on complex heuristics, and it effectively unifies the training objective with downstream decision-making quality. We conduct extensive experiments on various text retrieval benchmarks. The results demonstrate that when the training objective aligns with the evaluation setup, \alg{} yields remarkable in-domain performance improvement, with substantial out-of-domain generalization to some critical datasets employed in downstream question answering tasks. \footnote{Our data and model are publicly available at \url{https://huggingface.co/NeuralPGRank}.}

\end{abstract}

\section{Introduction}

Retrieving relevant factual information has become a fundamental component of modern language processing pipelines, as it grounds the decisions of the system and its users in factual sources. In particular, the retrieved text is often utilized by downstream application models to generate accurate outputs for various tasks, ranging from web search~\citep{Huang2013LearningDS}, question answering~\citep[]{Voorhees1999TheTQ, Chen2017ReadingWT, Karpukhin2020DensePR}, and open-ended generation~\citep{Lewis2020RetrievalAugmentedGF, Paranjape2021HindsightPT, Yu2022RetrievalaugmentedGA}.
This retrieval process not only acts as a knowledge base and reduces the search space for downstream models, but also can provide users with evidence to understand and validate the model's final output. Consequently, the quality of the retrieval system plays a pivotal role, significantly influencing the accuracy and completeness of any downstream decision making.

Recent research has seen a significant performance boost from incorporating pre-trained large language models into the retrieval policy \citep[e.g.,][]{Nogueira2019PassageRW, Lin2020PretrainedTF, Karpukhin2020DensePR}.
LLM-based text retrievers excel in contextualizing user queries and documents in natural language, often handling long-form or even conversational inputs. While these neural models generally outperform traditional count-based methods, training high-performing LLM-based retrieval policies presents several challenges.

The primary challenge lies in the complex nature of rankings as combinatorial objects, such that formulating efficient training objectives to enhance LLM-based retrieval functions becomes challenging due to the large number of potential rankings. Existing training methods thus commonly resort to optimizing pairwise preferences as an approximation. Unfortunately, these pairwise training objectives do not directly relate to the desired ranking metrics for retrieval, such as nDCG (Normalised Cumulative Discount Gain) or MRR (Mean Reciprocal Rate). To ameliorate this mismatch, most approaches rely on complex heuristics that are difficult to control, including the careful selection of specific hard negative examples~\citep{Xiong2020ApproximateNN}, employing a distillation paradigm~\citep{Qu2020RocketQAAO, Yang2020NeuralRF}, or adopting an iterative training-and-negative-refreshing approach~\citep{Sun2022ReduceCF}. 
As a result of these intertwined challenges, training a competitive-performing retrieval system is very difficult.

\begin{figure*}[!t]
    \centering
    \vspace{15pt}
    \includegraphics[width=\linewidth]{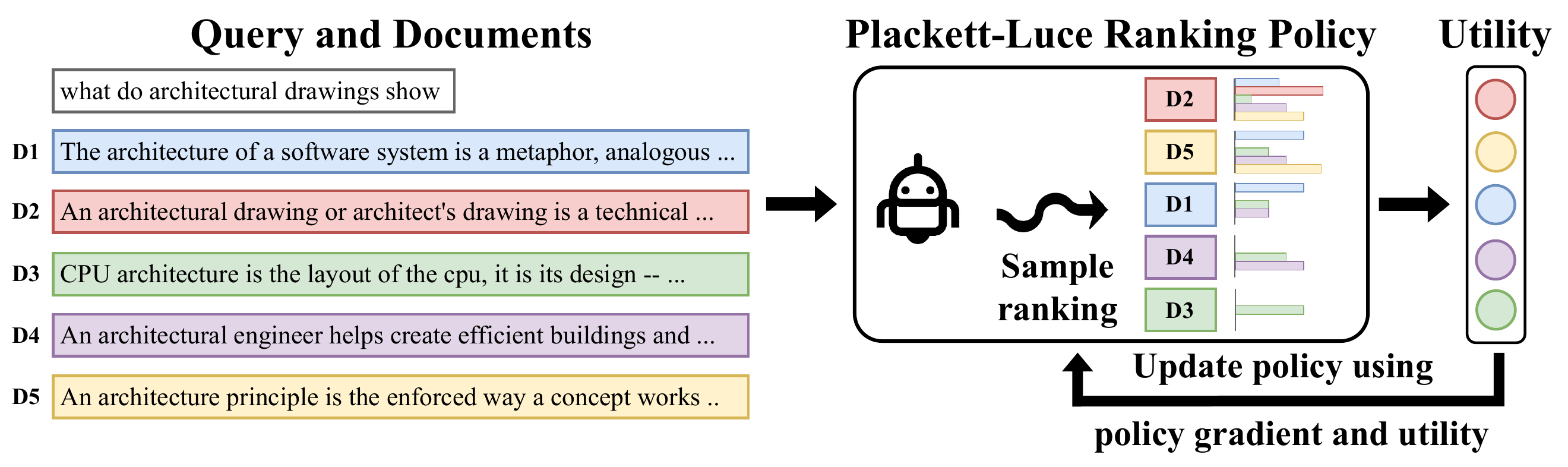} \vspace{-10pt}
    \caption{Illustration of our \alg{}. Given a query and a collection of documents, a Placket-Luce ranking policy samples ranking, receives utility, and gets updated using policy gradient and the received utility. Our method can directly optimize any ranking metric of interest as utility, and allows end-to-end training of any differential policy. Query and document examples are from MS MARCO dataset~\citep{Campos2016MSMA}.}
    \label{fig:main}
    \vspace{-10pt}
\end{figure*}

To overcome the above issues, we propose \alg{}, a rigorous and principled method that directly learns to rank through policy-gradient training. Our approach enables end-to-end training of any differentiable LLM-based retrieval model as a Plackett-Luce ranking policy. Moreover, our method can directly optimize any ranking metric of interest, effectively unifying the training objective with downstream application utility. This enables \alg{} to not only optimize standard ranking metrics like nDCG, but any application specific metric that evaluates the eventual output of the processing pipeline (e.g., BLEU score). \autoref{fig:main} illustrates the proposed \alg{} framework: given a query and a collection of documents, a Plackett-Luce ranking policy samples rankings, receives utility, and updates itself using policy gradients based on the received utility. By minimizing the need for complex heuristics in negative selection and utilization, as well as eliminating the requirement for additional supervision, our method successfully addresses the aforementioned challenges while establishing a principled bridge between training objectives and downstream utility of retrieval models.
\autoref{tab:model-comparison} compares the reliance of state-of-the-art retrieval models, including our \alg{}, on negative document mining and additional supervision (more details in \autoref{sec:exp-setup}).

We conduct extensive experiments employing our \alg{} with different models on various text retrieval benchmarks. We investigate the effectiveness of our method in both first-stage retrieval (i.e. searching over the entire document collection) and second-stage reranking (i.e. searching within a smaller candidate set per query). The results demonstrate a compelling trend: when the training objective aligns with the evaluation setup, specifically within the context of second-stage reranking, \alg{} exhibits remarkable in-domain performance improvement.
Furthermore, we find substantial out-of-domain generalization from MS MARCO~\citep{Campos2016MSMA} to some critical datasets employed in downstream question answering tasks, such as NaturalQuestions~\citep[NQ;][]{Kwiatkowski2019NaturalQA} and HotpotQA~\citep{Yang2018HotpotQAAD}. 
Overall, our method and findings pave the way for future research endeavors dedicated to developing highly effective retrieval-based LLM pipelines that are tailored for practical, real-world applications.

\begin{table*}[!t]
    \caption{Reliance of state-of-the-art comparison systems and our \alg{} on negative document mining and additional supervision. Each check denotes a heuristics used during training. Our method minimizes the reliance on the type of negative documents, and does not require any additional supervision from other models to improve retrieval performance.} 
    \vspace{5pt}
    \centering \small
    \setlength{\tabcolsep}{1.1pt}
    \begin{tabular}{lccccc}
        \toprule
        \textbf{Method} & \multicolumn{3}{c}{\textbf{Source of Negative Documents}}& \multicolumn{2}{c}{\textbf{Additional Supervision}} \\
          & In-Batch & BM25 & Dense Model & Cross-Encoder & Late Interaction Model \\
         \midrule
         SBERT~\citep{reimers-2019-sentence-bert}            & & {\color{redberry}\ding{51}} & {\color{redberry}\ding{51}}{\color{redberry}\ding{51}}{\color{redberry}\ding{51}} & {\color{redberry}\ding{51}} & \\
         TAS-B~\citep{Hofsttter2021EfficientlyTA}            & {\color{redberry}\ding{51}} & {\color{redberry}\ding{51}} & & {\color{redberry}\ding{51}} & {\color{redberry}\ding{51}}\\
         SPLADEv2~\citep{Formal2021SPLADEVS}         & & {\color{redberry}\ding{51}} & {\color{redberry}\ding{51}}{\color{redberry}\ding{51}} & {\color{redberry}\ding{51}} & \\
         \alg{} (Ours)  & & & \color{redberry}\ding{51} & & \\
        \bottomrule
    \end{tabular}
    \vspace{-10pt}
    \label{tab:model-comparison}
\end{table*}

\vspace{10pt}

\section{Background and Related Work}

Information retrieval (IR) is a class of tasks concerned with searching over a collection to find relevant information to the given query. We focus on text retrieval, where query refers to a user input in natural language and the collection is composed of text documents of arbitrary length. Text retrieval is a central sub-task in many knowledge-intensive NLP problems.

\paragraph{Text Retrieval}
In the text retrieval literature, retrieval models have evolved from classic count-based methods to recent learning-based neural models. 
Conventional count-based methods, such as TF-IDF or BM25~\citep{Robertson2009ThePR}, rely on counting query term occurrences in documents, and do not consider word ordering by treating text as a bag of words.  They suffer from issues like lexical mismatch, where relevant documents may not contain exact query terms~\citep{Berger2000BridgingTL}. Prior work has explored how to enhance these lexical retrieval methods with neural networks~\citep{Nogueira2019DocumentEB, Cheriton2019FromDT, Zhao2020SPARTAEO}.

Starting from Latent Semantic Analysis~\citep{Deerwester1990IndexingBL}, dense vector representations have been studied to improve text retrieval, with recently arising popularity of encoding the query and document as dense vectors~\citep{Yih2011LearningDP, Huang2013LearningDS,Gillick2018EndtoEndRI}.
The advent of powerful LLMs has allowed for developing neural models to replace lexical methods, which are often referred as dense models~\citep{Nogueira2019PassageRW, Karpukhin2020DensePR, Humeau2020PolyencodersAA}. Dense models are typically trained in a supervised manner to differentiate relevant documents from irrelevant ones given the query by assigning higher scores to query-relevant documents. 
Architectures of commonly-used dense models include bi-encoders (or dual-encoders) which encode query and document separately and compute a similarity score between query and document embeddings~\citep{Guo2020MultiReQAAC,Liang2020EmbeddingbasedZR,Karpukhin2020DensePR, Ma2021ZeroshotNP,Ni2021LargeDE}, cross-encoders which take the concatenation of query and document and output a numerical relevance score~\citep{Nogueira2019PassageRW}, and late interaction models which leverage token-level embeddings of query and document from a bi-encoder to compute the final relevance score~\citep{khattab2020colbert, Santhanam2021ColBERTv2EA}.

In large-scale text collections, sampling query-irrelevant documents (conventionally called negatives) is necessary for feasible training. 
Improving negative sampling to obtain a better selection of negatives (i.e. hard negatives) has been an active area of research, such as mining hard negatives from BM25~\citep{Xiong2020ApproximateNN}, or from stronger models~\citep{Qu2020RocketQAAO,Formal2021SPLADEVS}.
Another strategy to boost the performance of dense retrieval models is to employ the knowledge distillation paradigm~\citep{Qu2020RocketQAAO}, where a teacher model can provide query-dependent relevance scores of documents for the student retrieval model to learn from. 
While negative selection and distillation can improve the retrieval performance, they unfortunately require complex heuristics and convoluted training pipelines.
We propose a method that minimizes the reliance on intricate heuristics during training and requires no additional supervision as learning signals. Our method also closes the gap between training objective and evaluation metrics to improve not only the ranking in isolation, but also to directly optimize the overall pipeline performance.

\paragraph{Learning to Rank} 
Learning-to-rank (LTR) has a rich history in the field of IR. Our work falls under the category of LLM-based methods, and for a comprehensive survey of non-LLM-based LTR retrieval models, we refer readers to \cite{liu2009learning}. 

LTR methods used in multi-stage retrieval pipelines have attracted significant interest from both academia \citep{matveeva2006high, wang2011cascade, asadi2013effectiveness, chen2017efficient, mackenzie2018query, Nogueira2019PassageRW, khattab2020colbert, luan2021sparse, guo2022semantic} and industry \citep{amazon_blog}. Well-known product deployments of such systems include the Bing search engine \citep{pedersen2010query}, Alibaba's e-commerce search engine \citep{liu2017cascade}, and OpenAI's ChatGPT plugins \citep{chatgpt_plugin}. 
The common thread among these studies is the integration of retrieval and ranking systems to ultimately learn effective retrieval strategies.

Among the works in the LTR literature, two that are closely related to our \alg{} approach are \cite{singh2019policy} and \cite{oosterhuis2021computationally}, which use Plackett-Luce models to learn a ranking policy. Both approaches extend LTR policies to stochastic policies, allowing for the maximization of task-relevant utility while incorporating fairness constraints during the learning process. In this work, we extend such framework to the context of multi-stage LTR and retrieval pipelines using LLMs, effectively unifying the training objective and ranking evaluation, with additional variance reduction techniques and dense learning signals.

\section{Setting}
\label{sec:setting}

We focus on retrieval problems that involve integrating a text retrieval system into a larger language-processing pipeline. In these applications, user queries can be lengthy and intricate natural language descriptions, and the retrieved results are often used as input for downstream models, which further process them to generate outputs for the overall task. This introduces two requirements that go beyond the traditional retrieval application in search engines. Firstly, the retrieval system must be capable of comprehending complex textual queries, which motivates the utilization of powerful language models as part of the retrieval system. Secondly, it is crucial to optimize the entire set of retrieval results holistically, as the quality of the downstream answer depends on the collective set of retrieval results, rather than individual documents alone.

To address these requirements with a principled machine learning approach, we formalize the problem setting as follows. We assume a distribution $\mathcal{Q}$ from which queries are drawn.  Given a query $q$, we have a candidate set of $n$ documents $\mathbf{d}^{q} = \{d_1^{q}, d_2^{q}, \ldots, d_n^{q}\}$. 
Our goal is to train a ranking policy $\pi(r|q)$ that produces a ranking $r$ of the documents in the candidate set $\mathbf{d}^{q}$ given a query $q$. For full generality, we allow for stochastic ranking policies, which include deterministic ranking policies as a special case.

To evaluate the quality of a ranking $r$, we use an application-specific utility function $\Delta(r|q)$. This allows us to define the utility of a ranking policy $\pi$ for query $q$ as
\begin{equation}
    U(\pi | q) = \E_{r\sim\pi(\cdot | q)}\left[\Delta(r|q)\right].
\end{equation}
It is worth noting that $\Delta(r|q)$ can be any real-valued and bounded function that measures the quality of the entire ranking $r$ for query $q$. It does not necessarily need to decompose into relevance judgments of individual documents. For example, $\Delta(r|q)$ can be a function that quantifies the success of using ranking $r$ in a larger language processing pipeline for the overall task, enabling end-to-end optimization of the ranking policy $\pi$. Our learning objective is to learn a ranking policy $\pi$ that optimizes the expected utility over the query distribution $\mathcal{Q}$:
\begin{equation}
    \pi^{\star} = \argmax_{\pi\in\Pi}\E_{q\sim\mathcal{Q}}\left[U(\pi | q)\right]
    \label{eq:obj}
\end{equation}
where $\Pi$ represents the space of possible ranking policies.

To ensure compatibility with conventional training methods in the retrieval literature, our framework also covers the scenario where we have individual relevance judgments $\rel^{q}_i\in\{0, 1\}$ for each document in the candidate set, denoted as $\textbf{rel}^{q} = \{\rel^{q}_1, \rel^{q}_2, \ldots, \rel^{q}_n\}$. In this case,  $\Delta(r|q)$ could be a function like DCG (Cumulative Discount Gain), nDCG (Normalised DCG), 
MAP (Mean Average Precision), or MRR (Mean Reciprocal Rate). Specifically, for DCG,  we have $\Delta_{\text{DCG}}(r|q) = \sum_j \frac{u(r(j) | q)}{\log(1+j)}$ where $u(r(j) | q)$ is the utility of ranking document $d_j$ in the ordering $r$ for the query $q$.
Although our algorithm does not require individual relevance judgments, we focus on the commonly-used nDCG in order to compare with prior that relied on this ranking metric. 

\section{Method} \label{sec:method}

We present our method, \alg{}, which addresses the IR problem described in \autoref{sec:setting}.

\paragraph{Plackett-Luce Ranking Policy} 
To train our ranking policies, we consider the following functional form that is compatible with any score-based retrieval architecture. In particular, we define representation functions $\eta^q_\theta(q)$ and $\eta^d_\theta(d)$, which encode the query $q$ and the document $d$ into fixed-width vector representations, respectively. Additionally, we introduce a comparison function $\phi$ which takes these representations and computes a score:
\begin{equation*}
    s_\theta(q,d) \triangleq \phi(\eta^q_\theta(q), \eta^d_\theta(d))
\end{equation*} 
Under the Plackett-Luce model \citep{plackett1975analysis, luce1959individual}, we can define a ranking policy $\pi_\theta(r|q)$ based on the scores $s_\theta(q, d)$. The ranking policy is expressed as a product of softmax distributions:
\begin{equation}
    \pi_\theta(r|q) = \prod\limits_{i=1}^{n} \frac{\exp{s_\theta(q, d_{r(i)}})}{\sum_{j \in \{r(i),\ldots,r(n)\}} \exp{s_\theta(q, d_j)}}.
    \label{eq:plackett}
\end{equation}
Note that this family of Plackett-Luce ranking policies includes the policy that simply sorts the documents by their scores as a limiting case:
\begin{equation}
    \pi^{\text{sort}}_\theta(r|q) \triangleq \argsort_{d \in \mathbf{d}^{q}} s_\theta(q,d),
    \label{eq:plackett_sort}
\end{equation}
where $\argsort$ returns the indices that would sort the given array in descending order. 
In particular, the Plackett-Luce distribution converges to this sort-based policy when the scores are scaled by a factor $\tau$ with $\lim \tau \rightarrow \infty$.
One important distinction between Plackett-Luce policies and sort-based policies is that Plackett-Luce policies remain differentiable, which is a crucial property exploited by our training algorithm. Specifically, our policy $\pi_\theta(r|q)$ and its logarithm $\log\pi_\theta(r|q)$ are differentiable as long as our scoring model $s_\theta$ is differentiable.

\paragraph{REINFORCE} To solve the optimization problem defined in \autoref{eq:obj}, we propose a policy gradient approach based on insights from the LTR literature \citep{singh2019policy, oosterhuis2021computationally}.  Using the log-derivative trick pioneered by the REINFORCE algorithm \citep{williams1992simple}, we derive the policy gradient as follows:
\begin{align}
    \nabla_\theta U(\pi_\theta | q) &= \nabla_\theta \E_{r\sim\pi_\theta(\cdot|q)}\left[\Delta(r|q)\right] \nonumber\\
    &= \E_{r \sim \pi_\theta(\cdot | q)} \left[ \nabla_\theta \log \pi_\theta(r | q) \Delta (r|q) \right].
    \label{eq:pg}
\end{align}
\autoref{eq:pg} exploits the key insight that we can express the gradient of our utility as the expectation over rankings of the gradient of the log-probabilities (i.e. the policy gradient) from our ranking policy $\pi_\theta$. We can thus estimate \autoref{eq:pg} using Monte Carlo sampling, as detailed below.

\paragraph{Monte Carlo Sampling} A naive method for sampling rankings from the policy $\pi_\theta$ to estimate the gradient is to iteratively draw documents without replacement from the softmax distribution over the remaining documents in the candidate set until there are no more documents left. However, this process has a quadratic computational complexity with respect to the size $n$ of the candidate set. Instead, we can equivalently sample rankings more efficiently in $O(n \log(n))$ time by sampling an entire ranking using the Gumbel-Softmax distribution \citep{jang2016categorical} induced by our policy $\pi_\theta$.

Given a query $q$ and its respective candidate set $\mathbf{d}^q$, to sample an ordering $r$ of documents from our policy $\pi_\theta$, we first compute the scores $\pi_\theta(r(d) | q)$ for all documents $d$ in the candidate set, as defined in \autoref{eq:plackett}.
To sample from this induced distribution, we use the Gumbel-Softmax trick. For every document $d$ in the candidate set, we draw independent and identically distributed (i.i.d.) Gumbel samples from the Gumbel distribution $g_d\sim\text{Gumbel}(0, 1)$. Then, we calculate the softmax of the sum of the log scores and their corresponding Gumbel samples as follows:
\begin{equation*}
    x_d = \frac{\exp(\log \pi_\theta(r(d) | q) + g_d)}{\sum_{d\in{\mathbf{d}^q}}\exp(\log \pi_\theta(r(d) | q) + g_d)}
\end{equation*}
Finally, we sort the documents according to their $x_d$ values, resulting in the sampled ranking $r$. In practice, this sampling procedure allows us to sample rankings as fast as we can sort top-$K$ documents, resulting in a $O(n\log(n))$ runtime complexity.

\paragraph{Variance Reduction} To reduce the variance induced by our Monte Carlo estimates of the gradient, we incorporate a baseline into our objective. It is important to note that subtracting a baseline from the objective still provides an unbiased estimate of the gradient. Baselines are commonly employed in policy gradient methods to enhance the stability of the updates. In the case of \alg{}, we adopt the REINFORCE leave-one-out baseline \citep{kool2019buy}. The estimation of our policy gradient, based on $N$ Monte Carlo samples, can be expressed as follows:
\begin{align}
    \widehat{\nabla}_\theta U(\pi_\theta | q) &= \frac{1}{N}\sum_i \Big[ \nabla_\theta \log \pi_\theta(r_i | q) \Big(\Delta (r_i|q) - \frac{1}{N - 1}\sum_{j\neq i}\Delta (r_j|q)\Big)\Big].
    \label{eq:pg_baseline}
\end{align}
where $r_i$ is a sampled ranking and $q$ corresponds to a specific query. $\Delta (r_i|q)$ denotes the utility of the ranking $r_i$ for this query $q$. It subtracts the average utility for all other sampled rankings for this query. By including the leave-one-out baseline, we enhance the estimation of the policy gradient and mitigate the impact of high variance in the updates.

\paragraph{Utility} While our \alg{} applies to any utility function $\Delta (r|q)$, we focus on nDCG@10 in our experiments to be able to compare against conventional methods. Moreover, prior work~\cite[e.g.,][]{Wang2013ATA, Thakur2021BEIRAH} argues that nDCG offers both theoretical consistency and a practical balance suitable for both binary and graded sub-level relevance annotations. Following \cite{oosterhuis2021computationally}, we exploit the insight that the utility at rank $k$ only interacts with the probability of the partial ranking up to $k$, and the partial ranking after $k$ does not affect the utility before $k$. The estimation of our policy gradient is now:
\begin{align}
    \widehat{\nabla}_\theta U(\pi_\theta | q) &= \frac{1}{N}\sum_i \Big[\sum_k \nabla_\theta \log \pi_\theta(r_{i,k} | q, r_{i,1:k-1}) \nonumber\\
    & \Big(\text{nDCG} (r_{i,k:}|q, r_{i,1:k-1}) - \frac{1}{N - 1}\sum_{j\neq i}\text{nDCG} (r_{j,k:}|q, r_{i,1:k-1})\Big)\Big].
    \label{eq:pg_baseline_ndcg}
\end{align}

\section{Experimental Setup} \label{sec:exp-setup}
In numerous applications of text retrieval systems, a prevalent practice involves a two-stage procedure: initially, retrieving a limited set of candidate documents from the full collection (stage 1), and subsequently, re-ranking these initially retrieved candidate documents (stage 2). We investigate the effectiveness of our method in both stages by conducting extensive experiments with different models on various text retrieval benchmarks.

\paragraph{Data} We use MS MARCO~\citep{Campos2016MSMA}, a standard large-scale text retrieval dataset created from real user search queries using Bing search. 
We train on the train split of MS MARCO from the BEIR benchmark suite \citep{Thakur2021BEIRAH}. For tuning hyperparameters, we carve out a validation set of 7k examples from the training data. 

During training, we mimic the two-stage retrieval setup that an eventual production system would use. In particular, we generate candidate sets of 1k documents per query, composed of ground-truth relevant documents to the query and irrelevant documents. These irrelevant documents come from a stage 1 retriever, for which we typically use \textit{gtr-t5-xl} \citep{Ni2021LargeDE} model in this work. 

For in-domain evaluation, following prior work, we report performance on the dev set of MS MARCO. We also report out-of-domain zero-shot evaluation performance of our MS MARCO models on the subset of BEIR with readily available test sets.\footnote{We include the passage ranking task in TREC-DL 2019 \citep{Craswell2020OverviewOT}, a variant of
MS MARCO, as an out-of-domain evaluation set. This dataset is available as the test split of MS MARCO in BEIR.} BEIR contains several existing text retrieval datasets, ranging from Wikipedia, scientific, financial, and bio-medical domains.
\autoref{tab:data_stats} in \autoref{sec:data_stats} lists some details of our evaluation sets.

\paragraph{Evaluation Setup} We report nDCG@10~\citep[Normalised
Cumulative Discount Gain;][]{jarvelin2000ir} on each evaluation set by reranking the candidate set per query as a second-stage ranker (\autoref{sec:second-stage}), or over the full document collection as a first-stage retriever (\autoref{sec:first-stage}). In the second-stage ranking evaluation, our candidate set for each query comprises of the top-ranked documents obtained from \textit{gtr-t5-xl} as stage 1 ranker, which serve as irrelevant documents, as well as the ground-truth documents that are known to be relevant to the query. The inclusion of these ground-truth query-relevant documents within the candidate set aims to approximate the candidate set retrieved by an optimal first-stage retriever.

\paragraph{Comparison System} We compare to the following systems from prior work: 
\begin{itemize}
    \item \textbf{BM25}~\citep{Robertson2009ThePR} A bag-of-words retrieval approach that ranks a set of documents based on the occurrence of the query tokens in each document using TF-IDF.\footnote{\url{https://github.com/castorini/anserini}}

    \item \textbf{SBERT}~\citep{reimers-2019-sentence-bert} A bi-encoder, dense retrieval model using hard negatives mined by various systems. The objective combines a negative log likelihood loss and a MarginMSE loss, with reference margin scores generated by a cross-encoder model.\footnote{\url{https://huggingface.co/sentence-transformers/msmarco-distilbert-dot-v5} released on Hugging Face (updated on Jun 15, 2022).} 
    
    \item \textbf{TAS-B}~\citep{Hofsttter2021EfficientlyTA} A bi-encoder model trained with topic-aware queries and a balanced margin sampling technique, replying on dual supervision in a knowledge distillation paradigm. The loss function is a pairwise MarginMSE loss with both hard negatives from BM25 and in-batch negatives.\footnote{\url{https://huggingface.co/sentence-transformers/msmarco-distilbert-base-tas-b}}
    
    \item \textbf{SPLADEv2}~\citep{Formal2021SPLADEVS} A bi-encoder model trained by combining a regularization term to learn sparse representation and a MarginMSE loss with hard negatives. Hard negatives and the reference margin scores are generated with a dense model trained with distillation and a cross-encoder reranker.\footnote{\url{https://huggingface.co/naver/splade_v2_distil}}
\end{itemize}
Excluding BM25, the above supervised learning models are trained on MS MARCO with \textit{distilbert-base-uncased}~\citep{Sanh2019DistilBERTAD} as the initialization,  use dot product to compute query-document similarity, are in comparable scale, and represent the state-of-the-art performance of each approach. \autoref{tab:model-comparison} lists the reliance of these comparison systems and our method on the source of negative documents and additional supervision used during training. Our \alg{} minimizes the reliance on the type of negative documents, and does not require any additional supervision from other models to improve retrieval performance.

\paragraph{Ranking Policy} 
The representation model $\eta$ parameterizing our ranking policy is initialized with either SBERT or TAS-B as a warm start.\footnote{Our warmstart models exclude SPLADEv2, because our \alg{} method does not impose regularization to maintain its sparse representation learning.} Unless noted in our ablation experiments, we update the policy using our \alg{} (described in \autoref{sec:method}) for 6 epochs over the training data.

\paragraph{Implementation Detail} Our codebase is built upon BEIR~\citep{Thakur2021BEIRAH} and Sentence-Transformers~\citep{reimers-2019-sentence-bert}. 
We run all experiments on A6000 GPUs with 48GB of VRAM. Please see \autoref{app:implementation} for more implementation and hyperparameter details. %

\section{Experimental Result}
For models trained using our method, we present their results on each evaluation set both as a second-stage reranker over the candidate set (\autoref{sec:second-stage}) and as a first-stage retriever over the full document collection (\autoref{sec:first-stage}).

\subsection{Second-Stage Reranking} 
\label{sec:second-stage}

\begin{table*}[tb!] 
    \vspace{-20pt}
    \caption{Second-stage reranking: nDCG@10 in-domain results. * marks evaluations run by us using the publicly available checkpoint. Bold font represents the highest number per row, and underline shows the second highest. Light green color highlights the experiments where our \alg{} yields performance gain.} 
    \vspace{5pt}
    \centering \small
    \begin{tabular}{l r r r r r r r r r }
        \toprule
        \textbf{Dataset} & \multicolumn{3}{c}{\textbf{Comparison Systems}} & \multicolumn{2}{c}{\textbf{Ours: \alg{}}} \\
            & SBERT* & TAS-B* & SPLADEv2* & with SBERT & with TAS-B   \\ 
        \midrule
MS MARCO dev & 0.892 & 0.893 & 0.900 & \cellcolor{lightgreen}\textbf{0.987} & \cellcolor{lightgreen}\underline{0.982} \\
        \bottomrule
    \end{tabular}
    \vspace{-10pt}
    \label{tab:eval_rerank_in-domain}
\end{table*}

\begin{table*}[tb!] 
    \caption{Second-stage reranking: nDCG@10 results on out-of-domain datasets. * marks evaluations run by us using the publicly available checkpoint. Bold font represents the highest number per row, and underline shows the second highest. Light green color highlights the experiments where our \alg{} yields performance gain.} 
    \vspace{5pt}
    \centering \small
    \begin{tabular}{l r r r r r r r r r }
        \toprule
        \textbf{Dataset} & \multicolumn{3}{c}{\textbf{Comparison Systems}} & \multicolumn{2}{c}{\textbf{Ours: \alg{}}} \\
            & SBERT* & TAS-B* & SPLADEv2* & with SBERT & with TAS-B   \\ 
        \midrule
TREC-DL 2019 & \underline{0.743} & \textbf{0.749} & \textbf{0.749} & 0.742 & 0.741\\
TREC-COVID & \textbf{0.764} & 0.711 & \underline{0.731} & 0.690 & 0.630\\
NFCorpus & \underline{0.308} & 0.320 & \textbf{0.341} & 0.249 & 0.303\\
NQ & 0.836 & 0.836 & 0.854 & \cellcolor{lightgreen}\underline{0.869} & \cellcolor{lightgreen}\textbf{0.878}\\
HotpotQA & 0.747 & 0.785 & 0.834 & \cellcolor{lightgreen}\textbf{0.902} & \cellcolor{lightgreen}\underline{0.900}\\
FiQA-2018 & \underline{0.291} & 0.279 & \textbf{0.342} & 0.131 & 0.139\\
ArguAna & 0.351 & 0.479 & \textbf{0.480} & \cellcolor{lightgreen}\underline{0.354} & 0.443\\
Touché-2020 & \textbf{0.480} & 0.423 & \underline{0.460} & 0.363 & 0.361\\
Quora & 0.962 & \textbf{0.982} & \underline{0.967} & \cellcolor{lightgreen}0.963 & \cellcolor{lightgreen}\textbf{0.982}\\
DBPedia & 0.513 & 0.513 & \textbf{0.533} & \cellcolor{lightgreen}0.521 & \cellcolor{lightgreen}\underline{0.525}\\
SCIDOCS & 0.144 & \underline{0.151} & \textbf{0.163} & 0.108 & 0.136\\
FEVER & \textbf{0.931} & 0.911 & \underline{0.929} & 0.907 & \cellcolor{lightgreen}0.913\\
Climate-FEVER & \underline{0.442} & 0.433 & \textbf{0.444} & 0.438 & 0.383\\
SciFact & \underline{0.597} & 0.579 & \textbf{0.696} & 0.316 & 0.410\\
        \bottomrule
    \end{tabular}
    \vspace{-8pt}
    \label{tab:eval_rerank_out-of-domain}
\end{table*}

We report the performance of our trained models as a second-stage reranker, searching over a candidate set of 1k documents for each query.\footnote{BM25 is not compared in second-stage reranking, since it is commonly used only as a first-stage approach.} 

\paragraph{In-Domain Performance} 
\autoref{tab:eval_rerank_in-domain} presents the second-stage reranking performance of \alg{} using various warm-start policies, as measured by nDCG@10.  The results reveal that training with \alg{} leads to remarkable in-domain improvements over the warmstart SBERT and TAS-B models on MS MARCO dev set, with gains of +0.095 and +0.089 in nDCG@10, respectively. Notably, \alg{} achieves exceptional nDCG scores, approaching a perfect score of 1.0, not only for nDCG@10 (0.987 and 0.982) but also for nDCG@5 (0.986 and 0.981), nDCG@3 (0.985 and 0.978), and nDCG@1 (0.975 and 0.965).\footnote{We report nDCG@5, nDCG@3 and nDCG@1 of our method for second-stage reranking in \autoref{tab:eval_rerank_ndcg5}, \autoref{tab:eval_rerank_ndcg3} and \autoref{tab:eval_rerank_ndcg1} in the Appendix, including both in-domain and out-of-domain evaluation.}
In addition, the performance improvements after training with our method are more substantial when measured in nDCG@1, nDCG@3, and nDCG@5. For example, our method yields performance gains of 0.149 and 0.146 over the warm-start SBERT and TAS-B models in nDCG@1. Overall, these in-domain results consistently demonstrate that \alg{} provides remarkable in-domain performance improvements across various nDCG@k measures, with larger gains observed with smaller k values.

\paragraph{Out-of-Domain Generalization} 
\autoref{tab:eval_rerank_out-of-domain} shows the second-stage reranking performance of our method on out-of-domain datasets on the BEIR benchmark. In general, models trained with \alg{} demonstrate a level of generalization comparable to the baseline models.
Importantly, they notably outperform the baselines in the case of NaturalQuestions~\citep[NQ;][]{Kwiatkowski2019NaturalQA} and HotpotQA~\citep{Yang2018HotpotQAAD}, which are critical and widely-studied benchmarks in question answering (QA).
Our method achieves strong performance on these datasets, with scores of 0.869/0.878 on NQ and 0.902/0.900 on HotpotQA. Similar to the trend observed in in-domain results across different nNDCG@k measures, our method exhibits larger performance gains with smaller k values in out-of-domain generalization. Remarkably, on HotpotQA, our method using SBERT achieves an impressive nDCG@1 score of 0.974 (see \autoref{tab:eval_rerank_ndcg1} in the Appendix).
These observations are particularly promising, suggesting that our trained reranker exhibits substantial generalization to the QA domain. We plan to delve deeper into this aspect. Conversely, the datasets in which models trained using our method exhibit comparatively weaker generalization predominantly belong to the domains of science and finance -- we hope to investigate this further as well.

\paragraph{Ablation: Training Epochs}
We investigate how the duration of training impacts the performance of \alg{}, in both in-domain and out-of-domain scenarios. In \autoref{tab:eval_rerank_epoch_comparison} in the Appendix, we present the results for different training duration, specifically 0, 2, and 6 epochs. These results demonstrate that \alg{} achieves strong in-domain performance even with just 2 training epochs. However, there is a slight degradation in out-of-domain performance when the training duration is increased to 6 epochs. This suggests that \alg{} has the potential to enhance its out-of-domain generalization capabilities by carefully selecting the model to strike a balance between in-domain and out-of-domain performance.

\subsection{First-Stage Retrieval} \label{sec:first-stage}
In this section, we evaluate \alg{} in first-stage retrieval, which is to search over the entire document collection for each query. This task can be particularly challenging when dealing with extensive document collections, as is the case when searching through the 8.8 million documents in the MS MARCO dataset.

\autoref{tab:eval_full_in-domain} presents the results when we use \alg{} policies as first-stage retrievers, even though they were trained as a second-stage reranker. 
We find that training \alg{} for second-stage reranking is insufficient to match the performance of baseline systems when used as a first-stage retriever.\footnote{We observe the same finding in the out-of-domain evaluation, which is reported in \autoref{tab:eval_full} in the Appendix.} We conjecture that restricting training of \alg{} to a specific first-stage retriever creates blind-spots in the learned policies, leading to suboptimal performance in first-stage retrieval. To overcome this issue, we will investigate cutting-plane methods, which can enable efficient training even without candidate sets, and which have been shown to be highly effective (and provably convergent) for training other ranking and structured prediction methods \citep{Joachims/06a,Joachims/etal/09a}.

\begin{table*}[t!] 
    \vspace{-20pt}
    \caption{First-stage retrieval: nDCG@10 in-domain results. * marks evaluations run by us using the publicly available checkpoint. Bold font represents the highest number per row, and underline shows the second highest.}
    \vspace{5pt}
    \centering \small
    \begin{tabular}{l c c c c c c c c}
        \toprule
        \textbf{Dataset} & \multicolumn{4}{c}{\textbf{Comparison Systems}}  & \multicolumn{2}{c}{\textbf{Ours: \alg{}}} \\
            & BM25 & SBERT* & TAS-B* & SPLADEv2* & with SBERT & with TAS-B   \\ 
        \midrule
MS MARCO dev & 0.228 & \textbf{0.434} & 0.407 & \underline{0.433} & 0.416 & 0.401\\
        \bottomrule
    \end{tabular}
    \vspace{-8pt}
    \label{tab:eval_full_in-domain}
\end{table*}

\section{Conclusion}

In this work, we introduce Neural PG-RANK, a novel training algorithm designed to address challenges associated with training LLM-based retrieval models. As a rigorous approach that reduces the dependence on intricate heuristics and directly optimizes relevant ranking metrics, Neural PG-RANK has demonstrated its effectiveness when training objective aligns with evaluation setup — specifically, in the context of second-stage reranking — by exhibiting remarkable in-domain performance improvement and presenting substantial out-of-domain generalization to some critical datasets employed in downstream question answering. Our work establishes a principled bridge between training objectives and practical utility of the collective set of retrieved results, thereby paving the way for future research endeavors aimed at constructing highly effective retrieval-based LLM pipelines that are tailored for practical applications.

\subsubsection*{Acknowledgments}
This research was supported in part by NSFAwards IIS-1901168, IIS-2312865, OAC-2311521, and NSF project \#1901030. All content represents the opinion of the authors, which is not necessarily shared or endorsed by their respective employers and/or sponsors.
We thank Daniel D. Lee, Sasha Rush, Travers Rhodes, Chanwoo Chun, and Minh Nguyen for helpful discussions and support.

\newpage

\bibliography{neurips_2023}
\bibliographystyle{iclr2024_conference}
\newpage
\appendix
\label{sec:appendix}

\section{Dataset Statistics} \label{sec:data_stats}
\autoref{tab:data_stats} reports some details of the evaluation datasets in BEIR that we report performance on. Most evaluation sets have binary annotation of the document relevance given the query (i.e. either relevant or irrelevant to the query), while some datasets provide graded annotation of the document relevance into sub-levels --
a grade of 0 means irrelevant, and positive grades (e.g., 3-level annotation gives 1, 2, or 3 as relevance judgement) marks relevant document. 

\begin{table}[tb!] 
    \caption{Details of our evaluation sets (test set unless noted otherwise): source domain of documents (Domain), number of queries (\# Q), number of documents in the full collection, (\# D), average number of relevant documents per query (\# Rel. D/Q), and the type of relevance annotation (Annotation). 
    }
    \centering \small
    \begin{tabular}{l r r r r r r r r}
        \toprule
        \textbf{Dataset} & \textbf{Domain} & \textbf{\# Q} & \textbf{\# D} & \textbf{\# Rel. D/Q} & \textbf{Annotation}\\
        \midrule
MS MARCO dev & misc. & 6,980 & 8.8M & 1.1 & binary & \\
TREC-DL 2019 & misc. & 43 & 9.1k & 95.4 & 3-level & \\
TREC-COVID & bio-medical & 50 & 171.3k & 439.5 & 3-level & \\
NFCorpus & bio-medical & 323 & 3.6k & 38.2 & 3-level & \\
NQ & Wikipedia & 3,452 & 2.7M & 1.2 & binary & \\
HotpotQA & Wikipedia & 7.405 & 5.2M & 2.0 & binary & \\
FiQA-2018 & finance & 648 & 57.6k & 2.6 & binary & \\
ArguAna & misc. & 1,406 & 8.7k & 1.0 & binary & \\
Touché-2020 & misc. & 49 & 382.5k & 19.0 & 3-level & \\
Quora & Quora & 10,000 & 522.9k & 1.6 & binary & \\
DBPedia & Wikipedia & 400 & 4.6M & 38.2 & 3-level & \\
SCIDOCS & scientific & 1,000 & 25.7k & 4.9 & binary & \\
FEVER & Wikipedia & 6,666 & 5.4M & 1.2 & binary & \\
Climate-FEVER & Wikipedia & 1,535 & 5.4M & 3.0 & binary & \\
SciFact & scientific  & 300 & 5,2k & 1.1 & binary & \\
    \bottomrule
    \end{tabular} 
    \label{tab:data_stats}
\end{table}
 
\section{Implementation Detail}
\label{app:implementation}
\autoref{tbl:hparams} lists the hyperparameters used in our experiments. Note that we use the same training hyperparameters across all experiments with different warmstart models in our work.
\begin{table}[tb!]
\centering
    \caption{Hyperparameters used for \alg{}.}
   \begin{tabular}{ll}
       \toprule
       \textbf{Setting} & \textbf{Values} \\
       \midrule
       model   & [SBERT, TAS-B]\\
       \midrule
       \alg{} & epochs: 6 \\
              & batch size: 220 \\
              & learning rate: 1e-6 \\
              & entropy coeff: 0.01 \\
              & \# rankings sampled per epoch: 5000 \\
              & gumbel softmax temperature ($\tau$): 0.05 \\
              & similarity function: dot product \\ 
       \bottomrule
   \end{tabular}
   \vspace{0.5mm}
   
   \label{tbl:hparams}
\end{table}

\section{Performance Tables}
\paragraph{Second-Stage Reranking}
In addition to NDCG@10 reported 
in \autoref{sec:second-stage}, we report NDCG@1 in \autoref{tab:eval_rerank_ndcg1}, NDCG@3 in \autoref{tab:eval_rerank_ndcg3}, and NDCG@5 in \autoref{tab:eval_rerank_ndcg5} for the second-stage reranking performance of our models trained with \alg{}. \autoref{tab:eval_rerank_epoch_comparison} shows the performance at 0, 2, and 6 epochs of training. 0 epoch means the warmstart models.

\paragraph{First-Stage Retrieval}
\autoref{tab:eval_full} reports evaluation of our models trained on MS MARCO as a first-stage retriever on out-of-domain datasets in BEIR.

\begin{table*}[tb!] 
    \caption{Second-stage reranking: nDCG@5 results. * marks evaluations run by us using the publicly available checkpoint. $^\ddagger$ double dagger symbol means in-domain evaluation. Bold font represents the highest number per row, and underline shows the second highest. Light green color highlights the experiments where our \alg{} yields performance gain.} 
    \vspace{5pt}
    \centering \small
    \begin{tabular}{l r r r r r r r r r }
        \toprule
        \textbf{Dataset} & \multicolumn{3}{c}{\textbf{Comparison Systems}} & \multicolumn{2}{c}{\textbf{Ours: \alg{}}} \\
            & SBERT* & TAS-B* & SPLADEv2* & with SBERT & with TAS-B   \\ 
        \midrule
MS MARCO dev$^\ddagger$ & 0.884 & 0.884 & 0.892 & \cellcolor{lightgreen}\textbf{0.986} & \cellcolor{lightgreen}\underline{0.981}\\
TREC-DL 2019 & 0.753 & 0.765 & 0.757 & \cellcolor{lightgreen}\underline{0.767} & \cellcolor{lightgreen}\textbf{0.771}\\
TREC-COVID & \textbf{0.782} & 0.719 & \underline{0.758} & 0.717 & 0.659\\
NFCorpus & 0.338 & \underline{0.356} & \textbf{0.376} & 0.281 & 0.334\\
NQ & 0.822 & 0.822 & 0.842 & \cellcolor{lightgreen}\underline{0.860} & \cellcolor{lightgreen}\textbf{0.868}\\
HotpotQA & 0.730 & 0.769 & 0.819 & \cellcolor{lightgreen}\textbf{0.892} & \cellcolor{lightgreen}\underline{0.890}\\
FiQA-2018 & \underline{0.267} & 0.251 & \textbf{0.317} & 0.122 & 0.127\\
ArguAna & 0.291 & \textbf{0.435} & \underline{0.426} & \cellcolor{lightgreen}0.307 & 0.395\\
Touché-2020   & \textbf{0.526} & 0.439 & \underline{0.516} & 0.382 & 0.378\\
Quora & 0.959 & \textbf{0.981} & \underline{0.964} & \cellcolor{lightgreen}0.960 & \cellcolor{lightgreen}\textbf{0.981}\\
DBPedia & 0.517 & 0.513 & \textbf{0.529} & \cellcolor{lightgreen}\underline{0.524} & \cellcolor{lightgreen}0.514\\
SCIDOCS & 0.122 & \underline{0.127} & \textbf{0.134} & 0.092 & 0.114\\
FEVER & \textbf{0.925} & 0.904 & \underline{0.923} & 0.902 & \cellcolor{lightgreen}0.908\\
Climate-FEVER & 0.371 & 0.388 & \underline{0.398} & \cellcolor{lightgreen}\textbf{0.404} & 0.350\\
SciFact & \underline{0.575} & 0.558 & \textbf{0.674} & 0.279 & 0.379\\
        \bottomrule
    \end{tabular}
    \vspace{-10pt}
    \label{tab:eval_rerank_ndcg5}
\end{table*}

\begin{table*}[tb!] 
    \caption{Second-stage reranking: nDCG@3 results. * marks evaluations run by us using the publicly available checkpoint. $^\ddagger$ double dagger symbol means in-domain evaluation. Bold font represents the highest number per row, and underline shows the second highest. Light green color highlights the experiments where our \alg{} yields performance gain.} 
    \vspace{5pt}
    \centering \small
    \begin{tabular}{l r r r r r r r r r }
        \toprule
        \textbf{Dataset} & \multicolumn{3}{c}{\textbf{Comparison Systems}} & \multicolumn{2}{c}{\textbf{Ours: \alg{}}} \\
            & SBERT* & TAS-B* & SPLADEv2* & with SBERT & with TAS-B   \\ 
        \midrule
MS MARCO dev$^\ddagger$ & 0.872 & 0.872 & 0.881 & \cellcolor{lightgreen}\textbf{0.985} & \cellcolor{lightgreen}\underline{0.978}\\
TREC-DL 2019 & 0.748 & 0.764 & 0.758 & \cellcolor{lightgreen}\textbf{0.772} & \cellcolor{lightgreen}\underline{0.770}\\
TREC-COVID & \textbf{0.810} & 0.745 & \underline{0.770} & 0.735 & 0.669\\
NFCorpus & 0.364 & \underline{0.385} & \textbf{0.405} & 0.305 & 0.364\\
NQ & 0.804 & 0.806 & 0.821 & \cellcolor{lightgreen}\underline{0.846} & \cellcolor{lightgreen}\textbf{0.857}\\
HotpotQA & 0.712 & 0.749 & 0.799 & \cellcolor{lightgreen}\textbf{0.878} & \cellcolor{lightgreen}\underline{0.875}\\
FiQA-2018 & \underline{0.260} & 0.244 & \textbf{0.302} & 0.123 & 0.124\\
ArguAna & 0.245 & \textbf{0.385} & \underline{0.368} & \cellcolor{lightgreen}0.268 & 0.349\\
Touché-2020   & \textbf{0.549} & 0.467 & \underline{0.540} & 0.404 & 0.418\\
Quora & 0.955 & 0.979 & 0.960 & \cellcolor{lightgreen}0.956 & \cellcolor{lightgreen}\textbf{0.979}\\
DBPedia & \textbf{0.539} & 0.526 & 0.533 & \cellcolor{lightgreen}\textbf{0.539} & \cellcolor{lightgreen}\underline{0.528}\\
SCIDOCS & 0.140 & \underline{0.151} & \textbf{0.152} & 0.108 & 0.133\\
FEVER & \textbf{0.921} & 0.898 & \underline{0.918} & 0.895 & \cellcolor{lightgreen}0.902\\
Climate-FEVER & 0.350 & 0.369 & \underline{0.379} & \cellcolor{lightgreen}\textbf{0.401} & 0.346\\
SciFact & \underline{0.563} & 0.534 & \textbf{0.662} & 0.260 & 0.353\\
        \bottomrule
    \end{tabular}
    \vspace{-10pt}
    \label{tab:eval_rerank_ndcg3}
\end{table*}

\begin{table*}[tb!] 
    \caption{Second-stage reranking: nDCG@1 results. * marks evaluations run by us using the publicly available checkpoint. $^\ddagger$ double dagger symbol means in-domain evaluation. Bold font represents the highest number per row, and underline shows the second highest. Light green color highlights the experiments where our \alg{} yields performance gain.} 
    \vspace{5pt}
    \centering \small
    \begin{tabular}{l r r r r r r r r r }
        \toprule
        \textbf{Dataset} & \multicolumn{3}{c}{\textbf{Comparison Systems}} & \multicolumn{2}{c}{\textbf{Ours: \alg{}}} \\
            & SBERT* & TAS-B* & SPLADEv2* & with SBERT & with TAS-B   \\ 
        \midrule
MS MARCO dev$^\ddagger$ & 0.826 & 0.819 & 0.830 & \cellcolor{lightgreen}\textbf{0.975} & \cellcolor{lightgreen}\underline{0.965}\\
TREC-DL 2019 & 0.771 & 0.764 & \underline{0.795} & \cellcolor{lightgreen}\textbf{0.802} & 0.744\\
TREC-COVID & \textbf{0.810} & 0.740 & \underline{0.770} & \underline{0.770} & 0.700\\
NFCorpus & 0.406 & \underline{0.438} & \textbf{0.460} & 0.344 & 0.410\\
NQ & 0.758 & 0.752 & 0.770 & \cellcolor{lightgreen}\underline{0.815} & \cellcolor{lightgreen}\textbf{0.822}\\
HotpotQA & 0.884 & 0.904 & \underline{0.941} & \cellcolor{lightgreen}\textbf{0.974} & \cellcolor{lightgreen}\textbf{0.974}\\
FiQA-2018 & \underline{0.286} & 0.265 & \textbf{0.329} & 0.150 & 0.140\\
ArguAna & 0.147 & \textbf{0.245} & \underline{0.237} & \cellcolor{lightgreen}0.171 & 0.233\\
Touché-2020   & \textbf{0.561} & \underline{0.510} & \textbf{0.561} & 0.449 & 0.439\\
Quora & 0.946 & \underline{0.975} & 0.952 & \cellcolor{lightgreen}0.950 & \cellcolor{lightgreen}\textbf{0.976}\\
DBPedia & \textbf{0.618} & 0.570 & 0.585 & \underline{0.604} & \cellcolor{lightgreen}0.583\\
SCIDOCS & \underline{0.182} & 0.187 & \textbf{0.196} & 0.142 & 0.176\\
FEVER & \textbf{0.928} & 0.889 & \underline{0.916} & 0.885 & \cellcolor{lightgreen}0.893\\
Climate-FEVER & 0.432 & 0.446 & 0.453 & \cellcolor{lightgreen}\textbf{0.536} & \cellcolor{lightgreen}\underline{0.463}\\
SciFact & \underline{0.473} & 0.470 & \textbf{0.603} & 0.217 & 0.283\\
        \bottomrule
    \end{tabular}
    \label{tab:eval_rerank_ndcg1}
\end{table*}

\begin{table*}[t!] 
    \caption{Second-stage reranking: nDCG@10 results of evaluating the warmstart model, the model after training for 2 epochs and after 6 epochs. $^\ddagger$ double dagger symbol means in-domain evaluation. Bold font represents the highest number per row, and underline shows the second highest. Light green color highlights the experiments where our \alg{} yields performance gain.}
    \vspace{5pt}
    \centering \small
    \begin{tabular}{l c c}
        \toprule
        \textbf{Dataset} & \multicolumn{2}{c}{\textbf{Performance of \alg{} at Epoch 0 $\rightarrow$ 2 $\rightarrow$ 6}} \\
            & with SBERT & with TAS-B \\
        \midrule
MS MARCO dev$^\ddagger$   & 0.892 $\rightarrow$ 0.982 $\rightarrow$ 0.987  & 0.893 $\rightarrow$ 0.963 $\rightarrow$ 0.982 \\
Avg. on other BEIR datasets & 0.579 $\rightarrow$ 0.546 $\rightarrow$ 0.539 & 0.582 $\rightarrow$ 0.573 $\rightarrow$ 0.553 \\
        \bottomrule
    \end{tabular}
    \vspace{-10pt}
    \label{tab:eval_rerank_epoch_comparison}
\end{table*}

\begin{table*}[t!] 
    \caption{First-stage retrieval: nDCG@10 results on out-of-domain datasets. * marks evaluations run by us using the publicly available checkpoint. Bold font represents the highest number per row, and underline shows the second highest. Light green color highlights the experiments where our \alg{} yields performance gain.}
    \vspace{5pt}
    \centering \small
    \begin{tabular}{l c c c c c c c c}
        \toprule
        \textbf{Dataset} & \multicolumn{4}{c}{\textbf{Comparison Systems}}  & \multicolumn{2}{c}{\textbf{Ours: \alg{}}} \\
            & BM25 & SBERT* & TAS-B* & SPLADEv2* & with SBERT & with TAS-B   \\ 
        \midrule
TREC-DL 2019 & 0.506 & 0.703 & \underline{0.723} & \textbf{0.729} & 0.703 & 0.710\\
TREC-COVID & 0.656 & \underline{0.664} & 0.487 & \textbf{0.710} & 0.446 & 0.346\\
NFCorpus & \underline{0.325} & 0.298 & 0.315 & \textbf{0.334} & 0.147 & 0.243\\
NQ & 0.329 & \underline{0.498} & 0.455 & \textbf{0.521} & 0.384 & 0.386\\
HotpotQA & \underline{0.603} & 0.587 & 0.581 & \textbf{0.684} & 0.500 & 0.465\\
FiQA-2018 & 0.236 & \underline{0.286} & 0.276 & \textbf{0.336} & 0.124 & 0.133\\
ArguAna & 0.315 & 0.349 & \textbf{0.479} & \textbf{0.479} & \cellcolor{lightgreen}0.353 & \underline{0.442}\\
Touché-2020 & \textbf{0.367} & 0.224 & 0.171 & \underline{0.272} & 0.129 & 0.110\\
Quora & 0.789 & 0.833 & 0.835 & \underline{0.838} & \cellcolor{lightgreen}\textbf{0.839} & 0.832\\
DBPedia & 0.313 & 0.375 & \underline{0.385} & \textbf{0.435} & 0.365 & 0.358\\
SCIDOCS & \textbf{0.158} & 0.141 & \underline{0.145} & \textbf{0.158} & 0.085 & 0.096\\
FEVER & 0.753 & \underline{0.774} & 0.678 & \textbf{0.786} & 0.358 & 0.341\\
Climate-FEVER & \underline{0.213} & \textbf{0.235} & 0.193 & \textbf{0.235} & 0.044 & 0.035\\
SciFact & \underline{0.665} & 0.595 & 0.575 & \textbf{0.693} & 0.264 & 0.369\\
        \bottomrule
    \end{tabular}
    \label{tab:eval_full}
\end{table*}

\end{document}